\documentclass[letterpaper, 10 pt, conference]{ieeeconf}  

\usepackage[english]{babel}
\usepackage{graphicx}
\usepackage{float}
\usepackage[linesnumbered,lined,ruled,vlined]{algorithm2e}
\usepackage{xcolor}
\usepackage{amsmath}
\let\labelindent\relax
\usepackage{enumitem}
\usepackage{makecell}
\usepackage{multirow}
\usepackage{todonotes}
\usepackage{siunitx}
\usepackage{cite}
\usepackage[colorlinks, citecolor=green]{hyperref}
\newcounter{todocounter}

\IEEEoverridecommandlockouts                              

\title{\LARGE \bf
Topological Mapping for Manhattan-like Repetitive Environments
}

\author{Sai Shubodh Puligilla$^{*}$, Satyajit Tourani$^{*}$, Tushar Vaidya$^{*}$, Udit Singh Parihar$^{*}$, \\ Ravi Kiran Sarvadevabhatla and K. Madhava Krishna%
\thanks{*Denotes authors with equal contribution}%
\thanks{This work was supported by Rapyuta Robotics. All authors except Ravi Kiran S. associated with Robotics Research Center, IIIT Hyderabad. Ravi Kiran S. associated with Centre for Visual Information Technology, IIIT Hyderabad.
        {\tt\small \{mkrishna, ravi.kiran\}@iiit.ac.in}}%
}
\begin{document}

\maketitle
\thispagestyle{empty}
\pagestyle{empty}

\begin{figure*}[ht]
\begin{center}
  \includegraphics[width=380px]{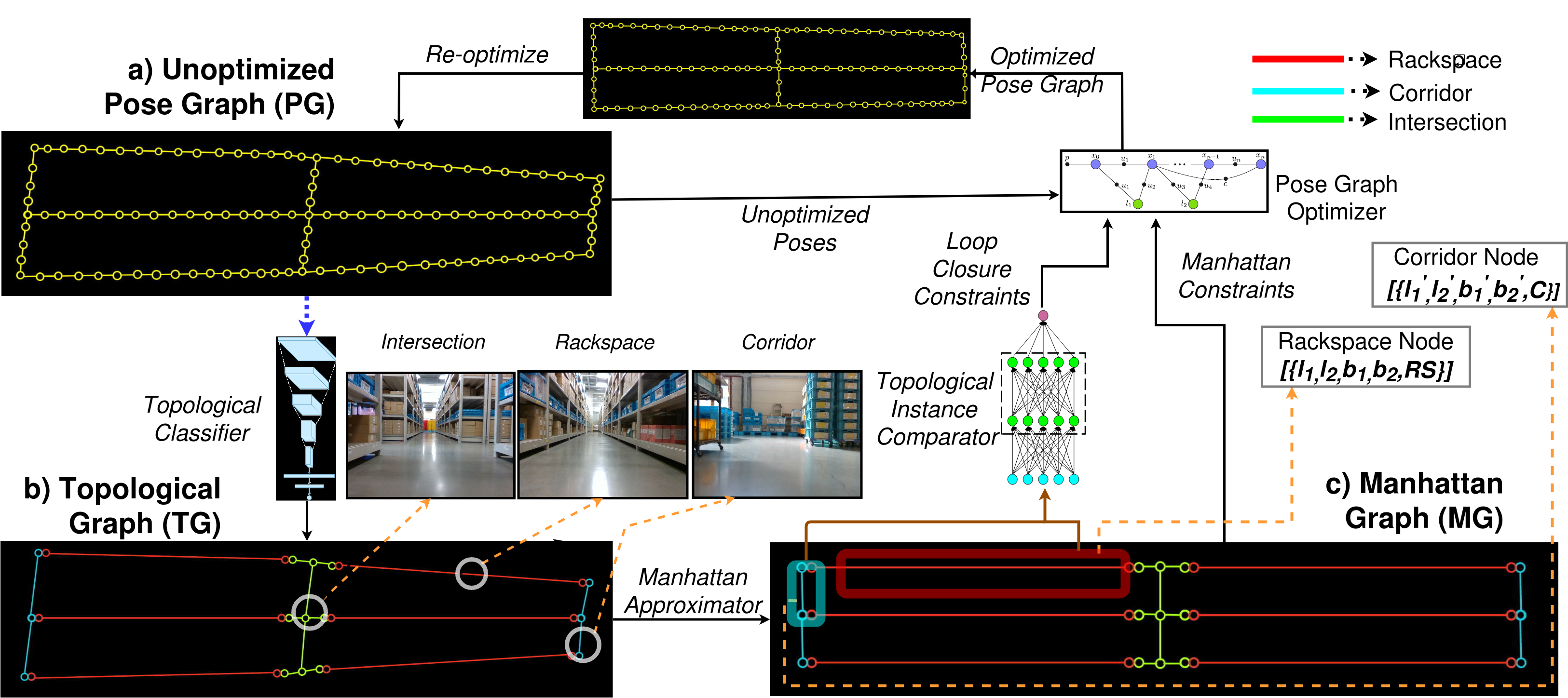}
  \caption{Overall formulation presented with Pose-graph as the input and CNN based classifier to classify regions. These classification labels are used to propose potentially similar regions in the pose-graph using Topological Instance Comparator which generates more constraints for the pose graph. Additionally, we generate topologically consistent graph called Manhattan Graph, which helps add more constraints in pose-graph eventually giving more relevant constraints to the pose-graph to increase accuracy. The blue dotted arrow shows the start of the pipeline and then follows a feedback like structure.}
  \label{fig:fig1_pipeline}
\end{center}
\end{figure*}
\begin{abstract}
We showcase a topological mapping framework for a challenging indoor warehouse setting. At the most abstract level, the warehouse is represented as a Topological Graph where the nodes of the graph represent a particular warehouse topological construct (e.g. rackspace, corridor) and the edges denote the existence of a path between two neighbouring nodes or topologies. At the intermediate level, the map is represented as a Manhattan Graph where the nodes and edges are characterized by Manhattan properties and as a Pose Graph at the lower-most level of detail. The topological constructs are learned via a Deep Convolutional Network while the relational properties between topological instances are learnt via a Siamese-style Neural Network. In the paper, we show that maintaining abstractions such as Topological Graph and Manhattan Graph help in recovering an accurate Pose Graph starting from a highly erroneous and unoptimized Pose Graph. We show how this is achieved by embedding topological and Manhattan relations as well as Manhattan Graph aided loop closure relations as constraints in the backend Pose Graph optimization framework. The recovery of near ground-truth Pose Graph on real-world indoor warehouse scenes vindicate the efficacy of the proposed framework.
\end{abstract}


\section{Introduction}
\label{sec:intro_sec}
This paper explores the role of topological understanding and the concomitant benefits of such an understanding to the SLAM framework. Figure-\ref{fig:fig1_pipeline} shows an erroneous Pose Graph $(PG)$ labelled `a', while the topological graph $TG$ is shown labelled as `b' in the same figure succinctly. Each node in the $TG$ is labelled by the Deep Convolutional Network. The $TG$ is converted to a Manhattan Graph $(MG)$ wherein the Manhattan properties of the nodes (length or width of the topology) and edges are gleaned from the $PG$. While the $MG$ facilitates seamless loop detection between a pair of Manhattan nodes, such relations when integrated with a back-end SLAM framework, enable the recovery of an optimized pose-graph and corresponding map. The crux of the paper lies in detailing the framework and its efficacy in challenging real world settings of two different warehouses.

There have been a number of works in this area and a detailed review of such methods can be seen in \cite{garcia2015vision}. Prominent and well cited amongst these include~\cite{ulrich2000appearance, sunderhauf2012switchable,agarwal2013robust, pronobis2006discriminative, ranganathan2006rao, kosecka2003qualitative}. Most of these methods are focused exclusively on vision based loop detection with invariant descriptors. Many relate to an individual image as a distinct topology of the scene without relating such nodes to a meta-level label such as a rackspace, corridor, intersection etc.
The classification task of place categorisation has been extensively performed on large datasets such as~\cite{Zhou2014LearningDF, zhou2017places, Sunderhauf_2016_place_categorization}. Works such as~\cite{Galindo2005MultihierarchicalSM, Zender2008ConceptualSpatial} and more recently,~\cite{armeni20193dscenegraphs, rosinol2020dsg} present hierarchical scene representations and how they are useful for various navigation tasks. However, neither of these exclusively tackle highly challenging repetitive environments such as warehouses.

\cite{Chen_2019_graph_localization_networks} shows the use of topological constructs for robot navigation in a simulated environment with the help of graph neural networks, where topological features are embedded into the nodes of a graph network. However, the method relies on the availability of noise-free sensor input, along with distinct topologies.
\cite{ranganathan2006bayesian} shows how a Bayesian inference over topologies can be performed to obtain more accurate topological maps. However, it does not entertain  notions of meta-level topological labels that go beyond an immediate lower-level topology restricted to the scene seen by the robot.

In this paper, we distinguish ourselves by portraying how higher level/meta level topological constructs that go beyond an immediate frame/scene and the relations that they enjoy amongst them percolate to a lower level pose-graph and elevate their metric relations. In fact, we recover close to ground truth floor plans from a highly disorganized map at the start. This is the essential contribution of the paper. In addition, the following constitute our contributions:
\begin{enumerate}
\item A deep convolutional network capable of learning warehouse topologies.
\item A Siamese Neural Network based relational classifier which resolves topological element ambiguity and helps achieve an accurate pose graph purely based on Topological relations. 
\item We showcase a backend SLAM framework that integrates loop closure relations from an intermediate level Manhattan Graph to the lowest level Pose Graph and elevate a disoriented unoptimized map to a structured optimized map which closely resembles the floor plan of the warehouse. Apart from the loop closure relations, the SLAM integrates other Manhattan relations to the pose graph. Ablation studies show the utility of both loop and Manhattan constraints as well as the superior performance of an incremental topological SLAM over a full batch topological SLAM. (Refer to Table \ref{tab: ate_comparison}.)
\item We also show how the two-way exchange between the $MG$ and $PG$ further improves the accuracy of the $PG$.
This two-way exchange between the various levels of representation is unique to this effort. Refer to the bottom two rows in Table~\ref{tab: ate_comparison}.
\end{enumerate}

Through the above formulation, the paper essentially exploits the Manhattan properties present in indoor warehouse scenes to perform $PG$ recoveries. Project page:  \href{https://github.com/Shubodh/ICRA2020}{https://github.com/Shubodh/ICRA2020}.

\section{Methodology}
Consider an unoptimized pose graph $PG$ represented by its nodes as $V_{pg}$ and edges as $E_{pg}$. The edge relations are of the following kinds:
\begin{itemize}
    \item Odometry relation between successive nodes.
    \item Loop closure relation between a pair of nodes.
    \item Manhattan relation between a pair of nodes.
\end{itemize} 

We obtain odometry relations from fused ICP and wheel odometry estimates which gives us the initial pose graph, which is highly erroneous. We then leverage the topological and Manhattan level awareness to generate the loop closure and Manhattan relations and use them for pose graph optimization to recover accurate graphs. This whole process is divided into 3 sub-sections:

\begin{enumerate}
    \item Topological categorization using a convolutional neural network classifier and its graph construction.
    \item Constructing Manhattan Graph from the obtained Topological Graph and predicting loop closure constraints using Multi-Layer Perceptron.
    \item Pose graph optimization using obtained Manhattan and loop closure constraints.
\end{enumerate}

Each part of the pipeline is described in each sub-section below, followed by experiments and results which are explained in the next section.

\subsection{Topological Categorization and Graph Construction}
\label{sec:semantic_place_cat}

Every node $V_{i} \in V_{pg} $ is associated with a topological label $L(V_{i})$ or $L_{i}$ where $L_{i} \in L=\{rackspace, corridor, transitions\}$ for a warehouse scene. To obtain these topological labels from visual data, we train a Convolutional Neural Network (CNN) configured for classification. The training data consists of RGB images resized to $224 \times 224$ and paired with topological node labels. For our warehouse setting, the labels are \texttt{Rackspace, Corridor, Intersection}:
    \begin{itemize}
        \item \texttt{Rackspace}: Location on path between two rackspaces
        \item \texttt{Corridor}: Location on the warehouse boundary path common to rackspaces
        \item \texttt{Intersection:} A transition location on the path 
    \end{itemize} 
    
Figure-\ref{fig:fig1_pipeline} entails examples of frames and their topological labels. Using PyTorch framework~\cite{NEURIPS2019_9015}, we train a ResNet-18~\cite{he2016deep} architecture pre-trained on ImageNet~\cite{imagenet_cvpr09} with its final layer replaced by a $3$-neuron fully connected layer, corresponding to the possible topological node labels. During training, we optimize the network to minimize cross-entropy loss. To account for class imbalance, we use class-weighted loss~\cite{johnson2019survey} with the following set of weights: \texttt{Rackspace}$:2.48$, \texttt{Corridor}$:2.16$, \texttt{Intersection}$:7.38$. The CNN is fine-tuned for $356$ epochs using Adam optimizer with a learning rate of $0.001$ for the pre-trained ResNet-18 layers and a learning rate of $0.005$ for the final layer weights. We stop training when the validation loss starts to increase. For training, we use $33{,}273$ images from two warehouses with a mini-batch size of $8$. To evaluate the trained network, we use $21{,}200$ images. The results are presented in the next section.

After obtaining the inferred labels from the CNN, we group together the adjacent nodes that share the same label. Thus, a node in Topological Graph $TG$ consists of two positions from the dense Pose Graph $PG$, i.e.  the starting and ending positions of that topology.  

\begin{figure}[!htbp]
            \includegraphics[width=8.8cm,height=6cm]{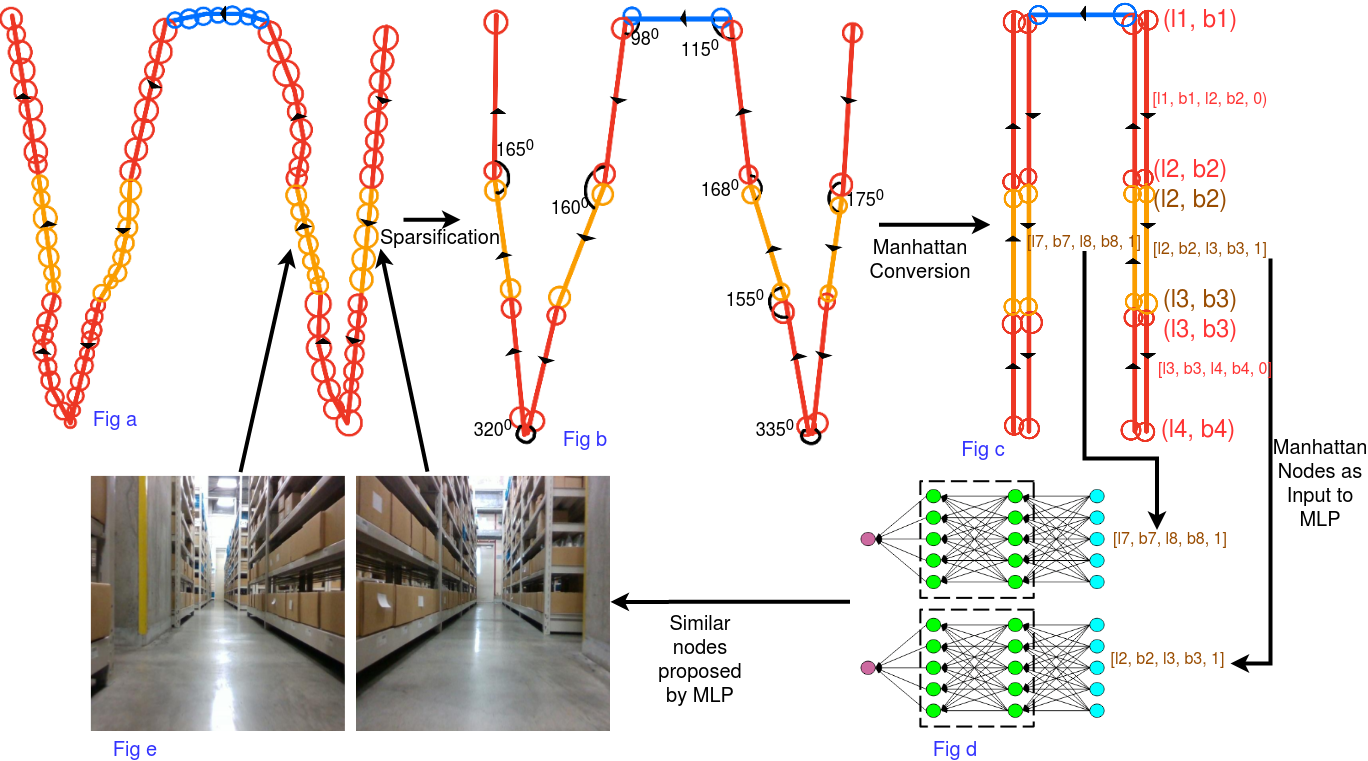}
            \caption{Fig.  a  shows  unoptimized  trajectory  with  dense  poses  along with  topological  labels.  Fig.  b  shows sparse  trajectory  obtained  by approximating each topological region with its starting and ending position. Notice that irregular robot trajectory inside some topological regions is also approximated to straight line. Fig. c shows Manhattan graph generated by approximating angles between two nodes to their nearest multiple of 90 degree. Fig. d is a siamese network which takes two nodes features as input and outputs similarity score between two nodes. Feature vector consists of starting and ending coordinates of a node along with node's topological label. Fig. e shows two images which are \ang{180} rotated but belongs to same intersection region as detected by siamese network. Arrows from Fig. e to Fig. a show corresponding locations of the two images in unoptimized pose graph.}
            \label{fig:manhattan_and_mlp}
\end{figure}


\subsection{Manhattan Graph Construction and Constraint Prediction using MLP}
\label{manhatten_construction}
We now explain how the Topological Graph $TG$ of the last section is converted to a Manhattan Graph, $MG$. We denote each node in the $MG$ as a meta-node, $M_{j} \in MG$, where $M_{j}$ corresponds to a collection $C$ i.e. $V_{i}, V_{i+1},.., V_{i+n}$ of $PG$ nodes, such that we write $V_{i} \in C(M_{j})$ and $L(M_{j}) = L(V_{i}) $ for every $PG$ node in the collection set $C$. A new meta-node is formed when there is a change in the label.

The pose graph nodes, their corresponding topology labels shown in the color denoting the label, the collection of such nodes that constitute a meta node also shown in the same color in the $MG$ are portrayed in figure~\ref{fig:manhattan_and_mlp}. 
	
The $MG$ relies on two essential measurements for its construction. 
\begin{itemize}
    \item The length $l$ of traversal or the length of topologies such as corridor or a rackspace.
    \item The angle $\phi$ made between two corridors/two rackspaces/rackspace and corridor via an intersection. 
\end{itemize} 

The length of the traversal is obtained by integrating fused odometry and ICP based transformations between two successive nodes of the Pose Graph that belong to the same meta node in the Manhattan Graph. The angle made as the robot moves from one topology (rackspace/corridor) to another (rackspace/corridor) via an intersection is estimated by fusing odometry and scan matching ICP measurements and integrating them over the traversal through the Intersection. This angle is binned to the closest multiple of $\pi/2$ as one of $\frac{-\pi}{2}, 0, \frac{\pi}{2}, \pi$. We use these sets of obtained lengths $l$ and angles $\phi$ along with the category of the meta node $M_{j}$ i.e. $L(M_{j})$ as attributes as input to a Siamese-style MLP neural network in order to determine if any two nodes in $MG$ are the same instance of a topological construct. In other words, the MLP determines if any two nodes in $MG$ correspond to the same topological area of the workspace. We chose to use MLPs for classifying the topologies instead of using a manual heuristic which might miss potential edge cases; however, note that any other classifiers such as k-nearest neighbours algorithm can be used. Figure 2 details the generation of Manhattan Graph and how its node features are fed into the MLP for loop closure detection. 

The training data for the MLP consists of what we have described as ``meta-nodes" above. Each meta-node is a tuple consisting of $\{X_{start}, Y_{start}, X_{end}, Y_{end}\}$. The four values of the tuple denote the starting and ending displacement co-ordinate of a particular node with respect to a global origin (global origin is the point from where the robot starts moving in the warehouse). $X_{start}$, $X_{end}$ and $Y_{start}$, $Y_{end}$ denote the displacement co-ordinates in the x and y-direction respectively. We create training data on the fly since we know the general structure of our warehouse and hence can create nodes synthetically using random numbers with similar lengths. The architecture is a Siamese network \cite{bromley1994signature} which consists of two hidden layers. We apply contrastive loss on the output obtained from the Siamese network to constrain semantically similar ``meta-node" representations to lie closer to each other. During inference, the MLP compares two nodes of the Manhattan Graph and predicts if the nodes correspond to the same topological instance. We base our approach on two strong assumptions:
        \begin{itemize}
        \item Each node comprises of one contiguous region of one particular category.
        \item Each node has displacement only in one direction. (Along x or y). 
        \end{itemize}


The classification that results from the MLP is particularly powerful due to its ability to classify two topological instances to be the same even when viewed from opposing viewpoints. This is shown in Figure-\ref{fig:manhattan_and_mlp} where the same topology is viewed from opposite viewpoint and have little in common. Yet the MLP's accurate classification of them to be the same instance becomes particularly useful for the Pose Graph optimization described in the next section. 

The MLP's non reliance on perceptual inputs also comes in handy for repetitive topologies. Warehouse scenes are often characterizes by repetitive structure and are prone to perceptual aliasing. The classification accuracy of the MLP is unaffected by such repetitiveness in the environment since it bypasses perceptual inputs. Yet the MLP does make use of perceptual inputs minimally in that it attempts to answer if the two nodes in the MG are the same instances only if the topological labels of the two nodes are predicted to be the same by the CNN.

\subsection{Pose Graph Optimization}


The Manhattan relation that exists between two nodes $M_{i},M_{j}$ and represented as $ R(M_{i},M_{j}) = <\Delta x_{ij}, \Delta y_{ij}, \Delta\theta_{ij}>$ serves as a Manhattan constraint between the nodes corresponding to $M_{i},M_{j}$ in the pose graph (in a manner consistent with the edge relations given in posegraph libraries such as G2O \cite{kummerle2011_g2o}, GT-SAM \cite{kaess2008isam}). $\Delta\theta_{ij}$ is typically $0^\circ$ or $180^\circ$ depending on whether the topology is being revisited with the same or opposing orientation.

The output of the MLP classifier is also used to invoke loop closure constraints. A pair of nodes classified to be the same topological construct by MLP corresponds to two sets of pose-graph nodes in the unoptimized graph belonging to the same area. Multiple loop closure relations are thus obtained between the pose-graph nodes of these two sets. Apart from these, there exist immediate Manhattan relations between two adjacent rackspaces or two adjacent corridors or a rackspace adjacent to a corridor mediated through an intersection. All such relations that exist in the Manhattan Graph as well as the loop closure relations percolate to the nodes in the PG as described further below.

In effect the optimizer solves for \cite{sunderhauf2011brief}:
\begin{align*}
    X^{*} = \underset{X}{argmax}\medspace P(X|U)
    = \underset{X}{argmax}\medspace \prod_{i}P(x_{i+1} | x_{i}, u_{i})\\ \times \underset{\textit{Loop Closure Constraints}}{\underbrace{\prod_{i \in C(M_{i}), j \in C(M_{j})}P(x_{j} | x_{i}, c_{ij})}}\\ \times {\underset{\textit{Manhattan Constraints}}{\underbrace{\prod_{i \in N(M_{i}), j \in N(M_{j})}P(x_{j} | x_{i}, m_{ij})}}} 
\end{align*}

where $P(X|U)$ is posterior probability of posegraph $X$ over set of constraints $U$, $x_{i}$ and $u_{i}$ are $i^{th}$ pose and controls of the robot. The loop closure relation $c_{ij}$ between nodes $i$ and $j$ is obtained using ICP. There are in principle $n(C(M_{i})) \times n(C(M_{j}))$ loop closure relations that are possible between topological constructs $M_{i}, M_{j}$ where $C(M_{i})$ is the collection set of Manhattan node $M_{i}$ as described before  and $n(P)$ is the cardinality of the set $P$. Whereas in practice we only sample a subset of such relations to constrain the graph.

Similarly, the graph is also constrained by Manhattan relations $m_{ij}$ that are invoked between the pose-graph nodes that constitute the sets $N(M_{i})$ and $N(M_{j})$ where $N(M_{i})$ and $N(M_{j})$ represent the Pose Graph nodes within the neighbourhood of $M_{i}$ and $M_{j}$. 

Typically $N(M_{j}) \subseteq C(M_{j})$. More formally, let $S$ be the set that enumerates all loop closure pairs discovered by the MLP over a Manhattan Graph MG. i.e $S = \{(M_{i},M_{j}), (M_{j},M_{k}), ..., (M_{p},M_{q})\} $, where each element of the set is a loop closure pair on the graph and $\{M_{i},M_{j},..,M_{q}\}$ are the nodes of the $MG$. Let $i$ be an iterator iterating over the element of $S$, $S(i) = (M_{p}, M_{q})$.  Let $LC$ be the set of all loop closure relations, $c_{ij}$ obtained for every $S(i) \in S$ by sampling from the $n(C(M_{i})) * n(C(M_{j}))$ number of loop closures possible for every $S(i) \in S$. Similarly, let $M$ be the set of all Manhattan relation $m_{ij}$ obtained for every $S(i) \in S$ from the neighbouring nodes in the unoptimized graph $N(M_{i}), N(M_{j})$ for every $S(i) = (M_{i},M_{j}) \in S$. Then


\begin{align*}
    X^{*} &= \underset{X}{argmax}\medspace \prod_{i}P(x_{i+1} | x_{i}, u_{i}) 
    \prod_{c_{ij} \in LC}P(x_{j} | x_{i}, c_{ij}) \\
    &{\prod_{m_{ij} \in M}P(x_{j} | x_{i}, m_{ij})}
\end{align*}

		
		

\section{Experimentation and Results}
\subsection{Topological Categorization in a Real Warehouse Setting}

\begin{figure}[!htbp]
            \includegraphics[width=8.8cm,height=6cm]{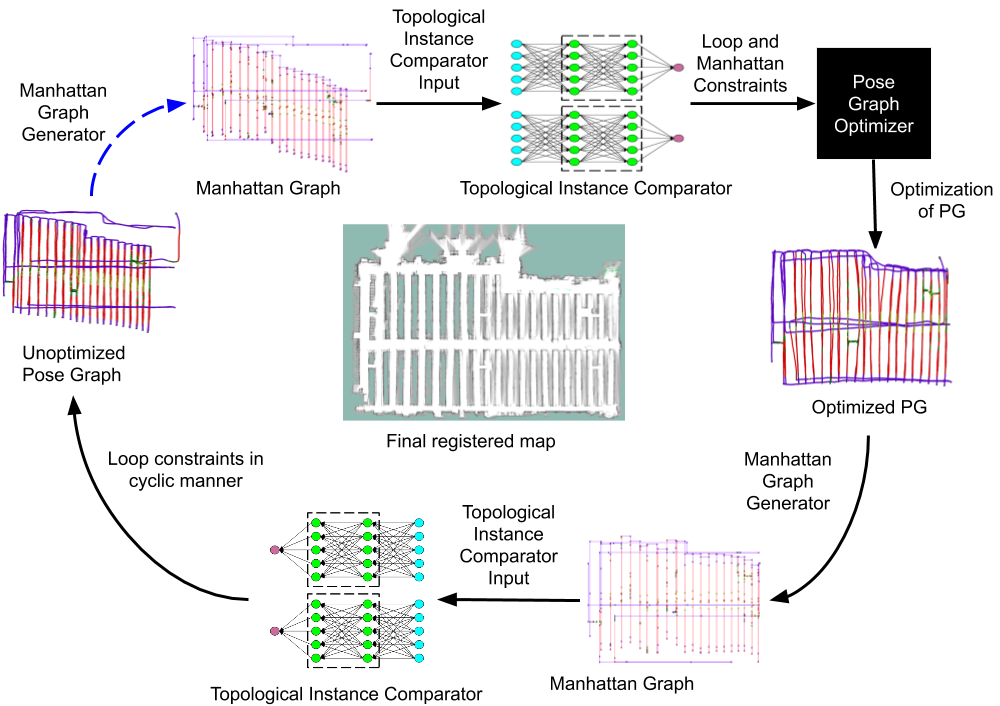}
            \caption{We start with the unoptimized Pose Graph with drift (shown by dotted-blue arrow) and obtain its Manhattan Graph. The nodes of the Manhattan graph are then passed to the Topological Instance Comparator which acts like a Topological Instance comparator and gives out node pairs which belong to the same topology. This is then passed to the Pose Graph Optimizer which gives us optimized Pose Graph. We then obtain the Manhattan graph of the optimized Pose Graph. The nodes of this graph are passed to the Topological Instance comparator which gives us improved(more accurate) loop pairs. This complete pipeline is done multiple times in a cyclic manner till convergence. We finally get a registered map which is shown in the center.}
            \label{fig:Posegraph Optimization pipeline}
\end{figure}

The performance of the topological node classification CNN (Section \ref{sec:semantic_place_cat}) can be viewed in Table \ref{CNN_accuracy_conf_results}. For the combined dataset, the network is able to classify the rackspace and corridor with very low false positives and false negatives with precision and recall more than $94 \%$ each. However, it is relatively difficult to classify the third class i.e. intersection as there is not much semantic consistency as the robot moves from one topology to another, which is reflected in the fact that the recall value is quite low, about $78 \% $. We explain how this inaccuracy affects the downstream modules in the Section \ref{sec:robust_analysis}.

\begin{table}[h]
\caption{CNN Classification Results.}
\label{CNN_accuracy_conf_results}
\begin{center}
\begin{tabular}{|c||c||c||c|}
\hline
\textbf{\thead{Warehouse \\ dataset}} & \textbf{Accuracy} \\
\hline
1\&2 &93.75\\
\hline
1&95.15\\
\hline
2&89.06\\
\hline
\end{tabular}
\quad
\begin{tabular}{|c||c||c|}
\hline
\multicolumn{3}{|c|}{\textbf{\thead{Metrics for Combined Data (1\&2)}}} \\
\hline
\textbf{\thead{Category}} & \textbf{Precision} & \textbf{Recall} \\
\hline
Rackspace & 94.2 & 96.3 \\
\hline
Corridor & 96.3 & 96.4 \\
\hline
Intersection & 85.6 & 78.1 \\
\hline
\end{tabular}
\end{center}
\end{table}

\subsection{Efficacy of Loop Closure Constraint Prediction using MLP}

            \begin{table}[h]
            \caption{MLP Results}
            \label{CNN_results_accuracy}
            \begin{center}
            \begin{tabular}{|c||c||c||c|}
            \hline
            \textbf{Network Type} & \textbf{Warehouse-1} &  \textbf{Warehouse-2}\\
            \hline
            Accuracy & 71.2 & 67.7\\
            \hline
            \end{tabular}
            \end{center}
            \end{table}

We showcase our pipeline on two different warehouses. There were two experiments performed. First, we  sample our training data according to the layout and length constraints of warehouse-1 and use the data-points of warehouse-1 as the lone testing data. In our second experiment, we train our MLP specifically according to the layout and length constraints of warehouse-2. 

The detection of nodes belonging to the same topology was observed to be accurate at the initial phase of the trajectory. The latter part of the trajectory was not accurate and had drift due to which the detection of node-pairs was observed to be inaccurate (Unoptimized Pose Graph shown in  Figure-\ref{fig:Posegraph Optimization pipeline}). We were able to improve the accuracy of the MLP and were able to generate accurate loop pairs by performing optimization on the Pose Graph in a cyclic fashion as shown in Figure-\ref{fig:Posegraph Optimization pipeline}.

We performed the experiments on both warehouses. The accuracy is calculated by checking for the percentage of the true node pairs. An accuracy of $71.2 \%$ and $67.7 \%$ was observed for the first and the second warehouse respectively. 

\subsection{Pose graph optimization Results}

The ablation study on the type of constraints have been done in five stages. The robustness of map recovery increases with each stage which reflects in Absolute Trajectory Error in Table \ref{tab: ate_comparison}. 
The experiments are conducted on four trajectories with varying lengths and starting deformation. Trajectories $W-1.1$ and $W-1.2$ are from the first warehouse and $W-2.1$ and $W-2.2$ are from the second warehouse in Table \ref{tab: ate_comparison}. Qualitatively, the initial trajectories are shown in first row of Figure \ref{fig:trajectory_recovery}.
\subsubsection{Stages of Map Recovery}
\begin{enumerate}[label=(\roman*)]
    \item \textit{Manhattan constraints: } Only manhattan constraints are used to optimize pose graph, PG. Constraints are extracted from manhattan graph, MG, between nodes proposed by MLP to be similar. 
    \item \textit{Loop Closure and Manhattan constraints: } Apart from Manhattan constraints, Loop Closure constraints as explained in section II-C are also used to constrain the PG. 
    \item \textit{Dense Proposals from MLP: } We consider nodes that have been classified to belong to the same instance with low confidence along with those classified to be the same with high confidence. This increases the number of constraints improving the optimization performance. The wrongly detected loops are filtered based on the loop closure (ICP) residual cost and do not make it to the optimization.
    \item \textit{Dense Proposals by MLP in Feedback Loop: } A feedback loop is invoked on the optimized PG from previous stage. A new MG is computed on the optimized PG, this manhattan graph, MG, is fed to MLP and sets of constraints are generated in a cyclic manner. This feedback mechanism leads to MLP performance improvement as shown in Table \ref{tab:my_label} and also helps in achieving very low Absolute Trajectory Error, ATE of 1.82 meters on four different maps from 11.57 meters in unoptimized map. This corresponds to the fourth of the contribution mentioned in the Section \ref{sec:intro_sec}.
    \item \textit{Incremental formulation: } Performing the feedback strategy from the previous stage in an incremental formulation in ISAM \cite{kaess2008isam} helps us to achieve the lowest ATE of 1.45 meters in our system. This confirms the robustness of our system to recover from highly unoptimized trajectories.
\end{enumerate}

\subsubsection{Qualitative Results}

We evaluate our system in two challenging real warehouse settings. The warehouse dimensions are $30m \times 50m$ and contains $21$ rack-spaces with intermediate corridors and intersections. All experiments  start with highly deformed trajectories. In all the cases, we were able to recover trajectories close to the groundtruth. Note that in our case, the ground-truth trajectory is the optimized map from the cartographer that has been confirmed with warehouse floor plan by our collaborators . These results are shown in Figure \ref{fig:trajectory_recovery}. The top row shows highly distorted pose graph trajectories while the middle row showcases the results of our optimization framework. The last row depicts the ground truth trajectories. The overall pipeline gets best illustrated with Figure \ref{fig:Posegraph Optimization pipeline}.

\subsection{Improving the performance of the state-of-the-art SLAM system: RTABMAP}


    

We compare our topological SLAM pipeline with the state-of-the-art SLAM system RTABMAP \cite{Rtabmap}, which is a highly modular library with the integration of various sensors like monocular camera, stereo camera, LiDAR, IMU and wheel odometry. When we evaluated RTABMAP on our warehouse dataset, we found that RTABMAP detects many False Positive loop closure constraints owing to similar looking corridors, and thus, incorrectly merges different parallel corridors into the same corridor. By incorporating our topological constraints in RTABMAP, we achieved better trajectory in terms of Absolute Trajectory Error, as shown in Table \ref{tab:rtabmap_vs_topo_ate}. This increase in performance is attributed to the non-reliance of MLP on individual frame-wise visual input but instead, it is utilizing the geometric structure of the topological representation. Hence, our topological constraints can be used to extend traditional Visual SLAM pipeline in highly repetitive Manhattan-like environments. Further qualitative results can be found in the Project page.

\begin{table}[!h]
\caption{ATE comparison of Topological SLAM with RTABMAP}
    \centering
    \begin{tabular}{|c|c|}
    \hline
        \textbf{\thead{RTABMAP}} & \textbf{\thead{RTABMAP + \\ Topological Constraints}}\\
    \hline
        4.45 & 3.36\\
    \hline
    \end{tabular}
    
    \label{tab:rtabmap_vs_topo_ate}
\end{table}

\subsection{Robustness Analysis}
\label{sec:robust_analysis}

We analyze the performance of the topological SLAM due to errors in topological classification due to the CNN and due to failure to detect loops by the MLP. Errors in topology classification manifest as loop detection in the MG. Therefore, the analysis is one of the robustness due to wrong loop detection  wherein both false positive and false negative cases are considered. The robustness to the pose graph optimization stems due from the following features:

\begin{enumerate}
    \item Residuals in the ICP estimated loop closure end up serving as priors to the element of dynamically scaled covariance matrix \cite{agarwal2013robust}, which serves as a robust kernel providing for backend topology recovery even when the number of wrong loop closures increase.
    \item An optimized PG feeds back to the MG and alleviates its error. The improved MG improve the loop detection performance of MLP, which percolate to the PG nodes and further improve its accuracy. Overtime this iterative exchange of information between the various representations improves the robustness of the PG backend.
\end{enumerate}

\begin{figure}[!h]
    \centering
    \includegraphics[width=5cm,height=3cm]{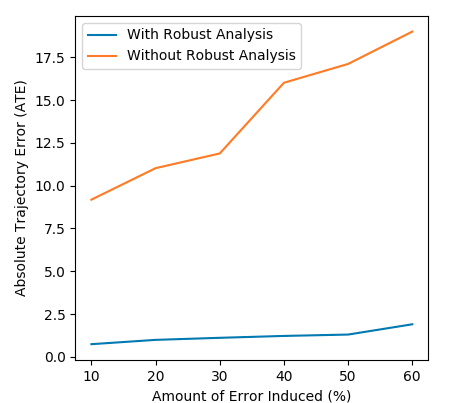}
    \caption{Effect of Robust pose-graph optimization with DCS}
    \label{fig:failure_analysis}
\end{figure}

To analyse the performance of our robust kernel exclusively in presence of the outliers, we synthetically introduced false positive (FP) and false negative (FN) loop closure pairs in the constraints for $PG$. In Figure \ref{fig:failure_analysis} X-axis represents percentage of loop closure pairs in the data-set having equal amount of FP and FN pairs. Y-axis represents ATE of TG with respect to the ground truth. The Figure \ref{fig:failure_analysis} depicts the performance of our framework due to errors occurred in topology classification and loop detection.

From the analysis of results in Figure \ref{fig:failure_analysis} it is evident that gradual increase in outliers can be tolerated by Robust kernel with DCS \cite{agarwal2013robust} as compared with non robust optimization techniques.

\begin{table}[!ht]
		\caption{Absolute Trajectory Error(ATE) \cite{sturm2012benchmark} for various pose-graphs with respect to ground-truth trajectories. LC stands for 'Loop Closure Constraints'}
		\centering
		\begin{tabular}{|p{2.5cm}|c|c|c|c|c|}
			\hline
			\multirow{2}{2.5cm}{\textbf{Method Type}} & 
			\multicolumn{1}{c|}{\textbf{W-2.1}} &
			\multicolumn{1}{c|}{\textbf{W-2.2}} &
			\multicolumn{1}{c|}{\textbf{W-1.1}} &
			\multicolumn{1}{c|}{\textbf{W-1.2}} &
			\multicolumn{1}{c|}{\textbf{Avg}} \\
			\cline{2-6}
			& \textbf{ATE} & \textbf{ATE}  & \textbf{ATE}  & \textbf{ATE}  &  \textbf{ATE}\\
			\hline
			Unoptimized & 4.7  & 7.5 &  16.3 & 17.8  & 11.57 \\ 
			\hline
			MLP Manhattan (G2O) & 3.42 & 2.85 &  4.5 & 7.4 &  4.54 \\
			\hline
			MLP  Manhattan + LC (G2O)& 3.09  & 2.7  & 3.9  & 1.67  & 2.84  \\ \hline
			Dense MLP  Manhattan + LC (G2O)  & 1.98  & 1.96 & 2.75  & 1.65  & 2.08  \\
			\hline
			Dense MLP  Manhattan + LC (In Feedback Loop) (G2O) & 1.67  & 1.8  & 2.21  & 1.6 &  1.82 \\ 
			\hline
			Dense MLP  Manhattan + LC (In Feedback Loop) (iSAM) & 1.6  & 0.98  & 1.02  & 2.2 & 1.45 \\
			\hline
		\end{tabular}
		
		\label{tab: ate_comparison}
\end{table}

\begin{table}[!h]
\caption{Effect of feedback mechanism on MLP Performance}
    \centering
    \begin{tabular}{|p{3cm}|c|c|c|}
    \hline
     & \textbf{\thead{True \\ Positive}} & \textbf{\thead{False \\ Positive}} & \textbf{Accuracy}\\
    \hline
        \thead{MLP} & 119 & 81 & 59.5\\
    \hline
      \thead{MLP in Feedback Loop }& 133 & 55 & 70.7\\
    \hline
    \end{tabular}
    
    \label{tab:my_label}
\end{table}

\begin{figure}[!h]
    \centering
    \includegraphics[width=230px,height=5.2cm]{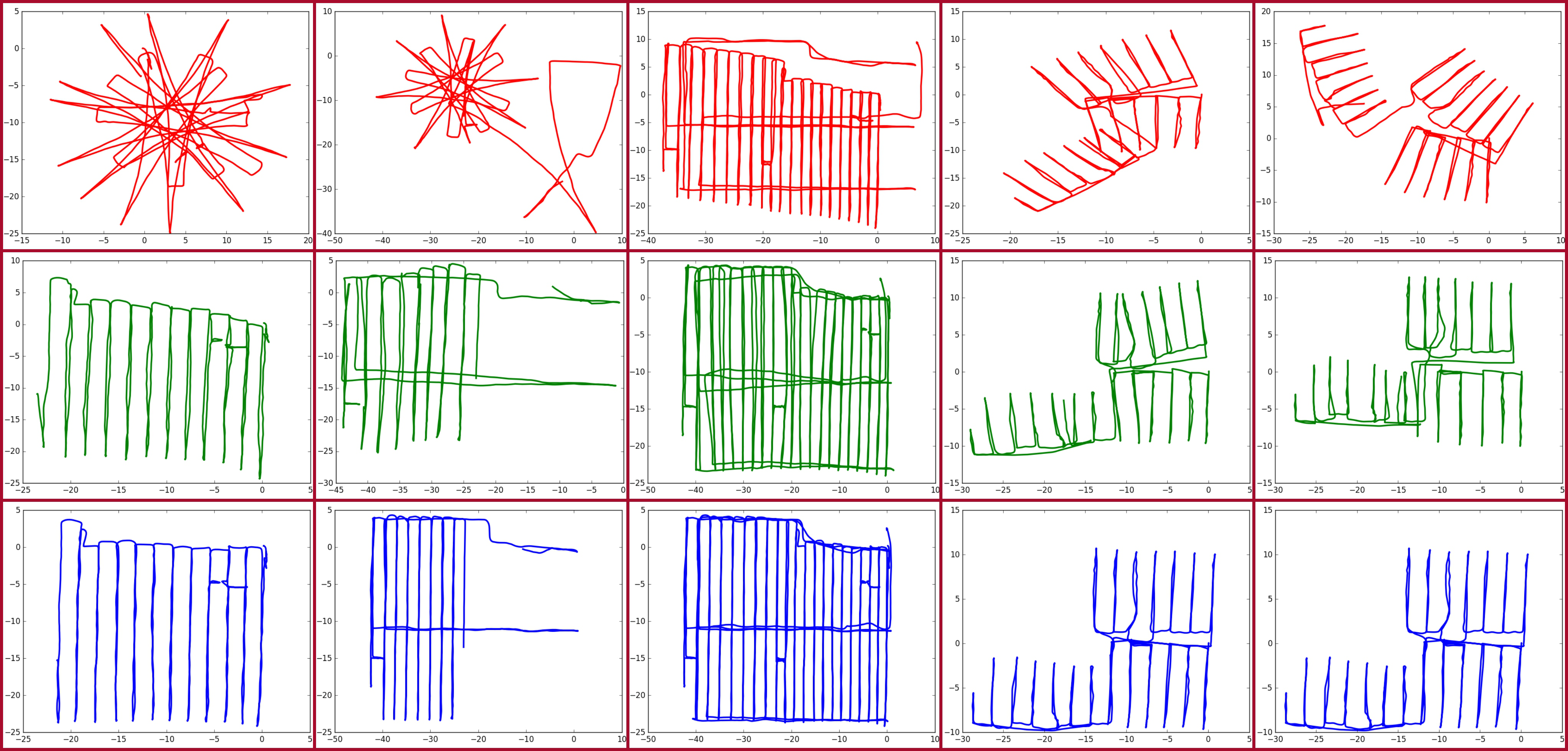}
    \caption{Top row shows unoptimized trajectories, middle row shows trajectories recovered using our pipeline and last row shows ground truth trajectories. Length in metres is shown along X and Y axes.}
    \label{fig:trajectory_recovery}
\end{figure}

\section{Conclusion}

This paper shows how higher level abstractions of an indoor workspace such as real warehouses can be used to effectively improve lower level backend modules of localization and mapping. Specifically we show how higher and intermediate level abstractions in the form of Topological Graph and Manhattan Graph can recover from backend pose graph optimization failures. Further by constant information exchange between the various levels of map abstractions we improve quantitatively the ATE by more than 87.4\% starting from very distorted pose graphs. We further show the method is robust to failures in the higher level representations such, which occurs when the Deep CNN architecture wrongly classifies a topological construct or when the Siamese style classifier wrongly detects or fails to detect loops in the Manhattan graph. The results shown are on two different real warehouse scenes over an area of around $30m \times 50m$, filled with many repetitive topologies in the form of corridor areas and rackspaces. Future results are intended to be shown on a variety of indoor topologies and office spaces such as for example those found in the Gibson environment \cite{xiazamirhe2018gibsonenv}.


\newpage
\bibliographystyle{unsrt}
\bibliography{citations}

\end{document}